\documentclass[letterpaper, 10 pt, conference]{ieeeconf}  

\IEEEoverridecommandlockouts                              

\usepackage{graphics} 
\usepackage{subfig}
\usepackage{epsfig} 
\usepackage{amsmath} 
\usepackage{mathtools}

\DeclarePairedDelimiter\floor{\lfloor}{\rfloor}
\title{\LARGE \bf
Cooperative LIDAR Object Detection via Feature Sharing in Deep Networks
}

\author{Ehsan Emad Marvasti$^{1}$, Arash Raftari$^{1}$, Amir Emad Marvasti$^{1}$, Yaser P. Fallah$^{1}$, Rui Guo$^{2}$, HongSheng Lu$^{2}$
\thanks{$^{1}$ Ehsan Emad Marvasti, Arash Raftari, Amir Emad Marvasti and Yaser P. Fallah are at University of Central Florida
        {\tt\small \{e\_emad,raftari,a\_emad\}@knights.ucf.edu, yaser.fallah@ucf.edu}}%
\thanks{$^{2}$ Rui Guo, Hongsheng Lu are principal researchers at Toyota
Motor North America R\&D InfoTech Labs, Mountain View, CA
94043 USA.{\tt\small \{ rguo,hlu\}@us.toyota-itc.com }}%
 }

\begin{document}

\maketitle
\thispagestyle{empty}
\pagestyle{empty}

\begin{abstract}
The recent advancements in communication and computational systems has led to significant improvement of situational awareness in connected and autonomous vehicles. Computationally efficient neural networks and high speed wireless vehicular networks have been some of the main contributors to this improvement. However, scalability and reliability issues caused by inherent limitations of sensory and communication systems are still challenging problems. In this paper, we aim to mitigate the effects of these limitations by introducing the concept of feature sharing for cooperative object detection (FS-COD). In our proposed approach, a better understanding of the environment is achieved by sharing partially processed data between cooperative vehicles while maintaining a balance between computation and communication load. This approach is different from current methods of map sharing, or sharing of raw data which are not scalable. The performance of the proposed approach is verified through experiments on Volony dataset. It is shown that the proposed approach has significant performance superiority over the conventional single-vehicle object detection approaches.

\end{abstract}

\begin{figure*}[!t]
\centering
\includegraphics[width=1\textwidth,trim={0mm 0mm 0mm 0mm},clip]{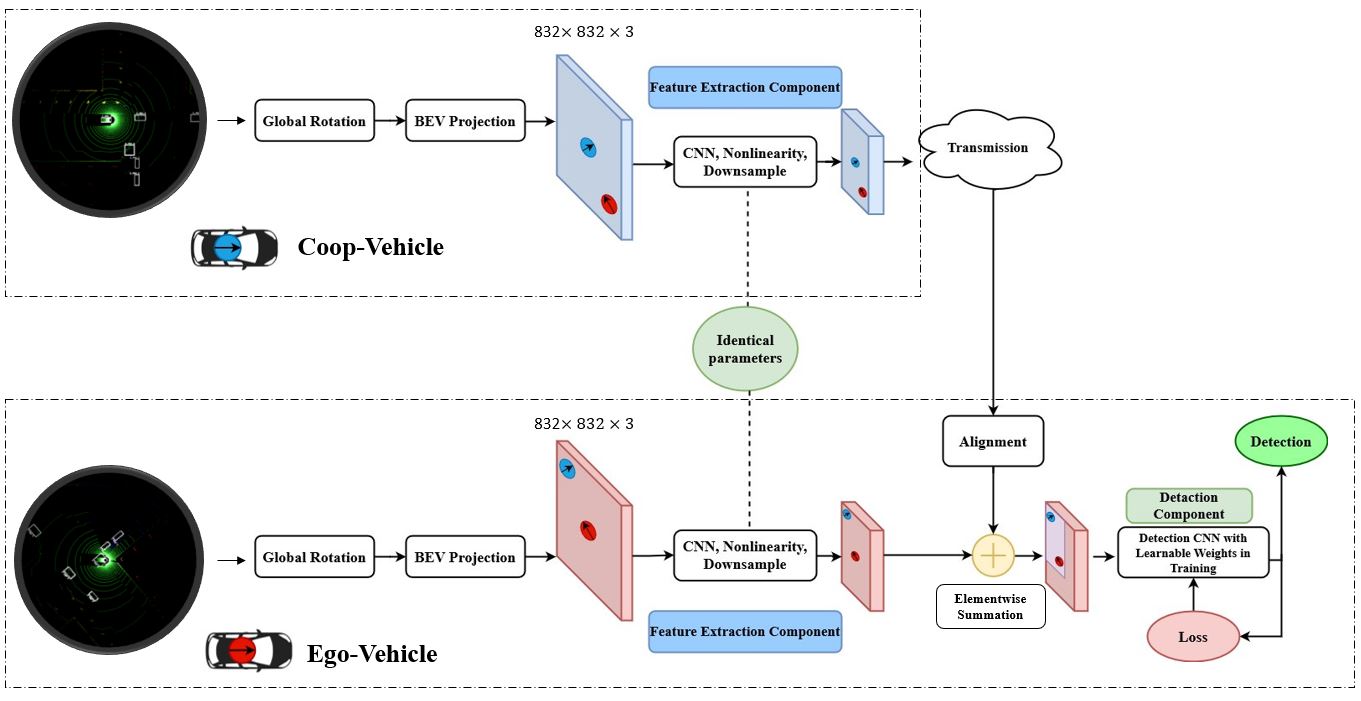}
\caption{The overview of feature sharing procedure. The cooperative vehicle transfers the LIDAR point-cloud to feature domain after an initial rotation alignment. After performing a translation transformation on the received feature-maps, the aligned received feature-maps are accumulated with the feature-map produced by receiver vehicle and fed through the object detection module}
\label{fig:overall}
\end{figure*}
\section{Introduction}
With the resurgence of inference methods made possible by computational systems advancements, the automotive industry is at a brink of a breakthrough in automation and safety improvements. The safety in automated vehicle systems is dependant on the robustness of the sensory and inference systems. Failure in each component can have catastrophic consequences; therefore, there have been constant efforts to improve the robustness of such systems. In many cases, improving the quality of sensory systems is costly; some devices also suffer from inherent limitations. For example, high-quality long-range LIDAR devices are prohibitively expensive for many low-cost manufacturing strategies. Moreover, even a high-quality LIDAR device inherently cannot overcome the inherent limitations of sensory units such as occlusion problem.

On the other hand, recent communication advancements for high speed wireless vehicular networks offer a new opportunity to overcome the quality and cost issues of sensory devices. Communication offers a possibility for cooperation among different vehicles equipped with sensors, possibly achieving synergistic gains in sensing performance. However, the limitations of wireless communication technologies have to be considered for designing cooperative methods. Vehicle to Everything (V2X) networks provide the possibility of information sharing amongst connected and autonomous vehicles (CAV), particularly, in a local area network. Using wireless links, vehicles can cooperate to aid each other in understanding their environment and ultimately improving their safety and efficiency. This paper aims at the cooperative cognition aspect of CAVs and attempts to improve a vehicle's situational awareness through communication.

In general, a vehicle's incomplete observation from the environment is due to either sensor obstruction, or resolution loss at larger distances leading to false or uncertain inference. In such situations, vehicles can collaborate at different levels to improve their view of the environment.

In vehicular academic and industrial  community, two methods have been proposed and vastly studied for improving vehicle's understanding of their environment: (1) sharing fully processed information or detection results and (2) sharing raw sensory data.
In \cite{8280503}, a map sharing technique combined with a content control strategy was proposed to extend 3D maps of LIDAR based data and enhance position tracking performance. An adaptive communication strategy was proposed for exchanging fully processed data (detection results) between participating entities to improve situational awareness by \cite{6953213}. In \cite{8891157}, a graph matching scheme amongst cooperative vehicles was introduced to build a 3D representation of the environment. The method efficiently maps the graph of objects detected by participating vehicles and infers occluded objects. Similarly, in \cite{8500388}, a scheme was proposed to increase the object detection confidence by accumulating the collaborative vehicles' perceptions (fully processed data) and constructing a global view of the scene. The aforementioned efforts take the approach of method 1 by sharing fully processed results. However, there are several other efforts, in particular from a sensor fusion perspective, that take the approach of method 2. In \cite{8403778}, the communication capacity requirement of collaborative sensing and benefits of exchanging sensors' raw information between participants in terms of sensing redundancy and range has been studied. Finally, performance improvement of collaborative sensing by deploying sensing and communication equipment at roadside infrastructure has been assessed in \cite{8417769}. 

While, in theory, sharing of sensory data is expected to yield the highest performance gain (e.g. in detection applications); the communication cost of sharing raw data is significant. On the other hand, the sharing of fully processed results will require much less communication load, but at the expense of lower performance. The option of sharing partially processed data, a new concept that is proposed in this paper, strikes a balance between performance and communication cost. 

In this paper, we present a novel approach of cooperative cognition by sharing partially processed data amongst cooperative vehicles. The partially processed data are the features derived from an intermediate layer of a deep neural network. It will be demonstrated how the concept of feature sharing is developed into a solution for cooperative object detection using data from LIDAR sensors. The results show that our proposed approach significantly improves the performance of object detection while keeping the required communication capacity low compared to sharing raw information methods. 

The rest of this paper is organized as follows. In section \ref{Section:Proposed Feature Sharing Cooperative Object Detection (FS-COD))}, the overall framework is introduced in details and its main building blocks are explained. In section \ref{Section:Experiments and Results}, details of dataset, the baseline method and the results of our experiments are provided. Finally, section \ref{Section:Concluding Remarks and Future work} concludes this work.

\section{Proposed Feature Sharing Cooperative Object Detection (FS-COD)}
\label{Section:Proposed Feature Sharing Cooperative Object Detection (FS-COD))}
 In this section, we propose a decentralized framework to improve object detection via feature-sharing between cooperative vehicles equipped with LIDAR.
Recent advances in deep learning has led to the development of object detection methods using Convolutional Neural Networks (CNN). 
YOLO \cite{Redmon_2016_CVPR} is among fast and reliable methods that gained attention due to their high performance and relatively low computational complexity which are suitable for real-time applications \cite{li2016vehicle,feng2018towards,beltran2018birdnet,10.1007/978-3-319-46448-0_2}. These methods are mainly distinguished by the optimization loss function, network architecture and input representations. In our work we have designed a CNN object detector by adapting the network architecture and loss function presented in \cite{simony2018complex} and \cite{Redmon_2016_CVPR} respectively.
However, other CNN based object detection methods can also be utilized in our scheme with some minor modifications.

In vehicular domain, target objects, e.g. vehicles and pedestrians, typically lie on a surface such as the road or side walk. Given this assumption, bird-eye view(BEV) projections of point-clouds have gained popularity in the field of vehicular object detection\cite{Yang_2018_CVPR,kim2019enhanced,yang2018pixor}.
Although BEV projection causes information loss, it has some merits in our particular application. BEV projection method will significantly reduce computational cost as opposed to volumetric methods\cite{arnold2019survey}, making it a suitable choice for real-time applications.
In addition, in BEV images, the size of the observed object remains constant and does not change with respect to its distance from the sensory unit similar to volumetric point-cloud methods. Hence, we have modified Complex-YOLO CNN backbone structure to exploit this characteristic and enhance the performance of the detection system in our setup. The modification details is provided in section \ref{Baseline: Single Vehicle Object Detection}.

Although these modifications can enhance the object detection performance in single-vehicle object detection setup, the challenges of detecting caused by non-line-of-sight and partial occlusion still exist. Concepts such as collective perception messages have been proposed to address the aforementioned issue; however, lack of consensus in inferences of cooperative vehicles might arise as a consequence. FS-COD is proposed as a solution to partial occlusion, sensor range limitation and lack of consensus challenges.

\begin{figure*}[!pt]
     \subfloat[]{%
       \includegraphics[width=.30\textwidth]{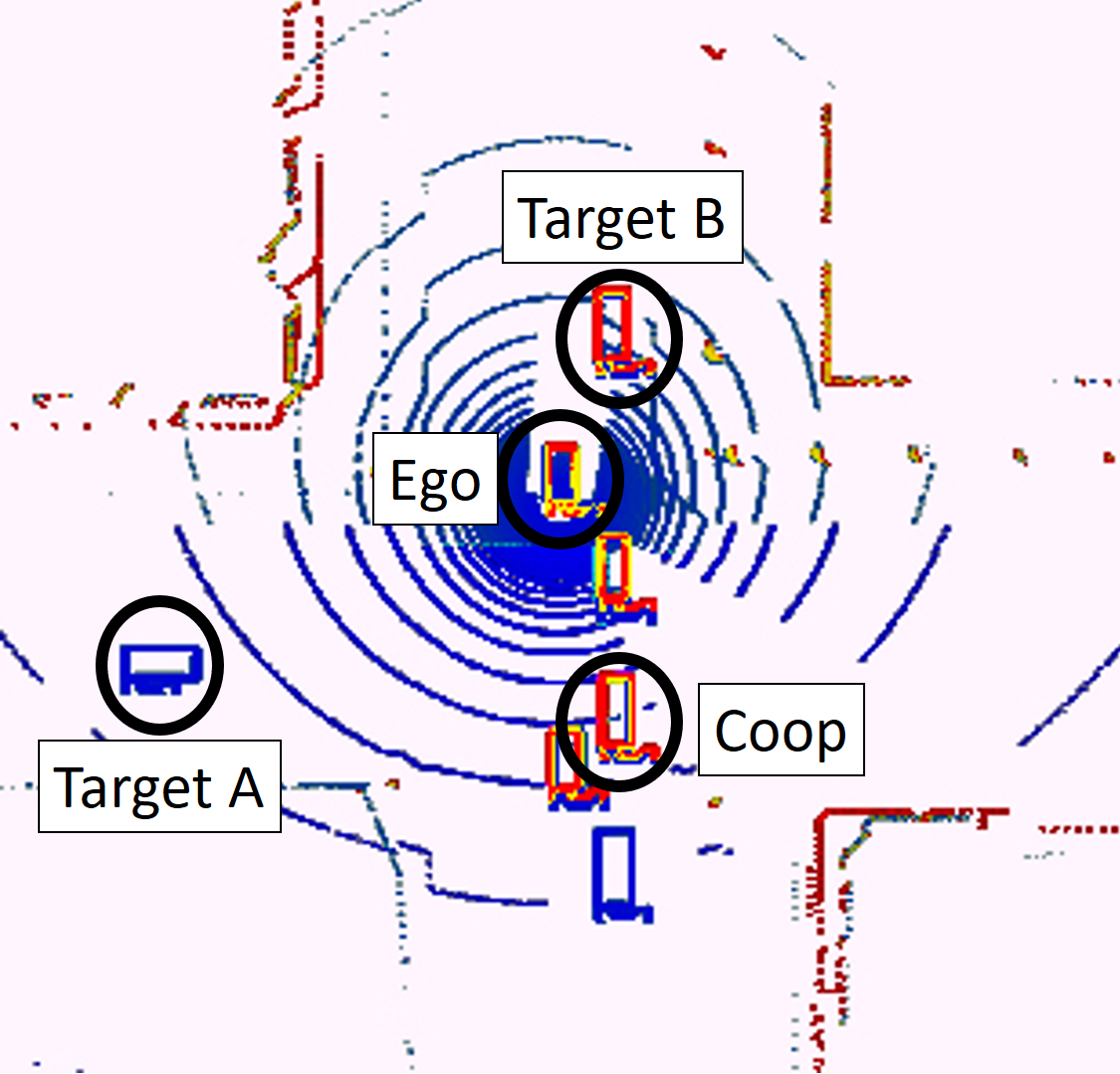}
     }
     \qquad
     \subfloat[]{%
       \includegraphics[width=.30\textwidth]{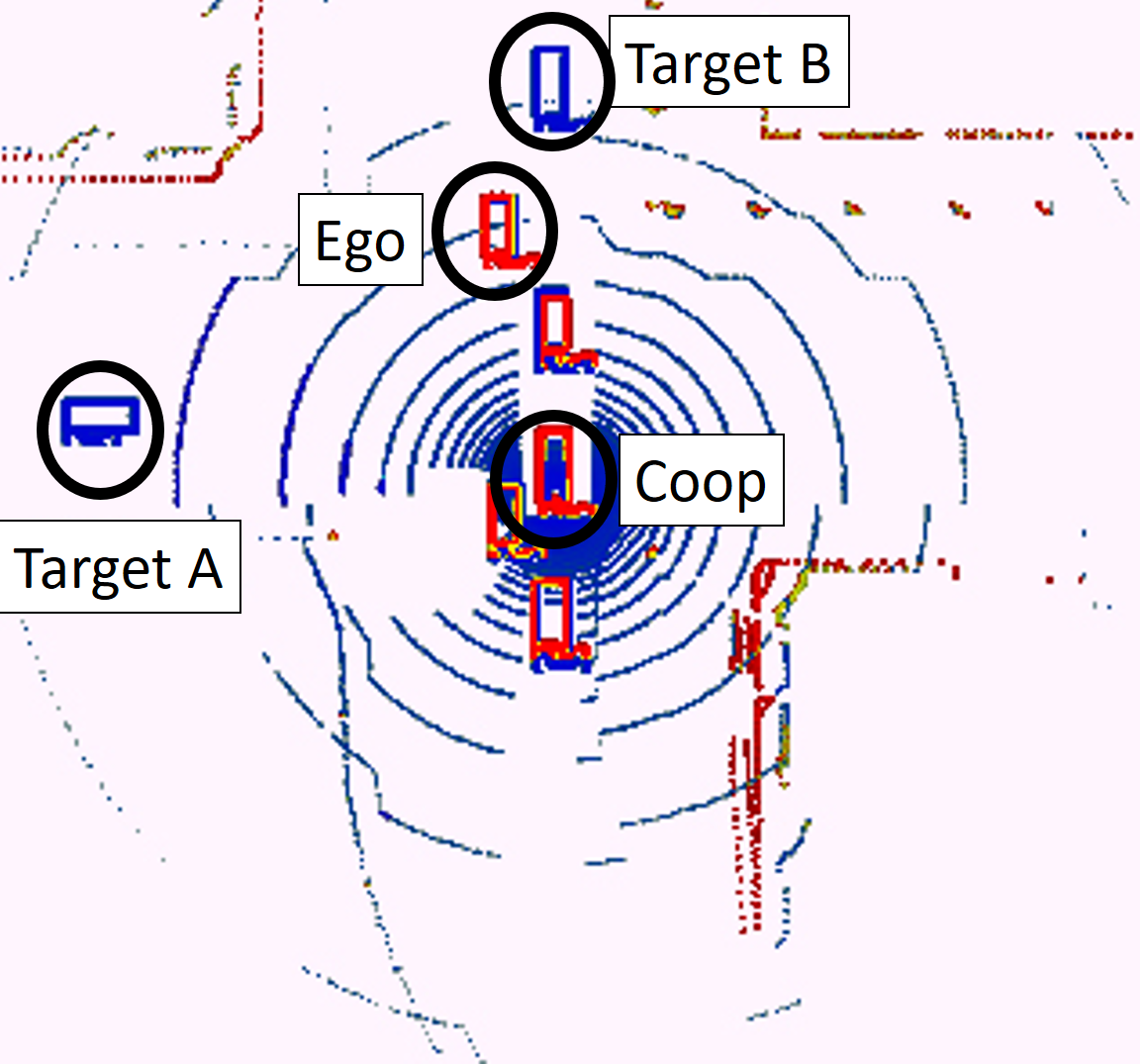}
     }
     \qquad
     \subfloat[]{%
       \includegraphics[width=.30\textwidth]{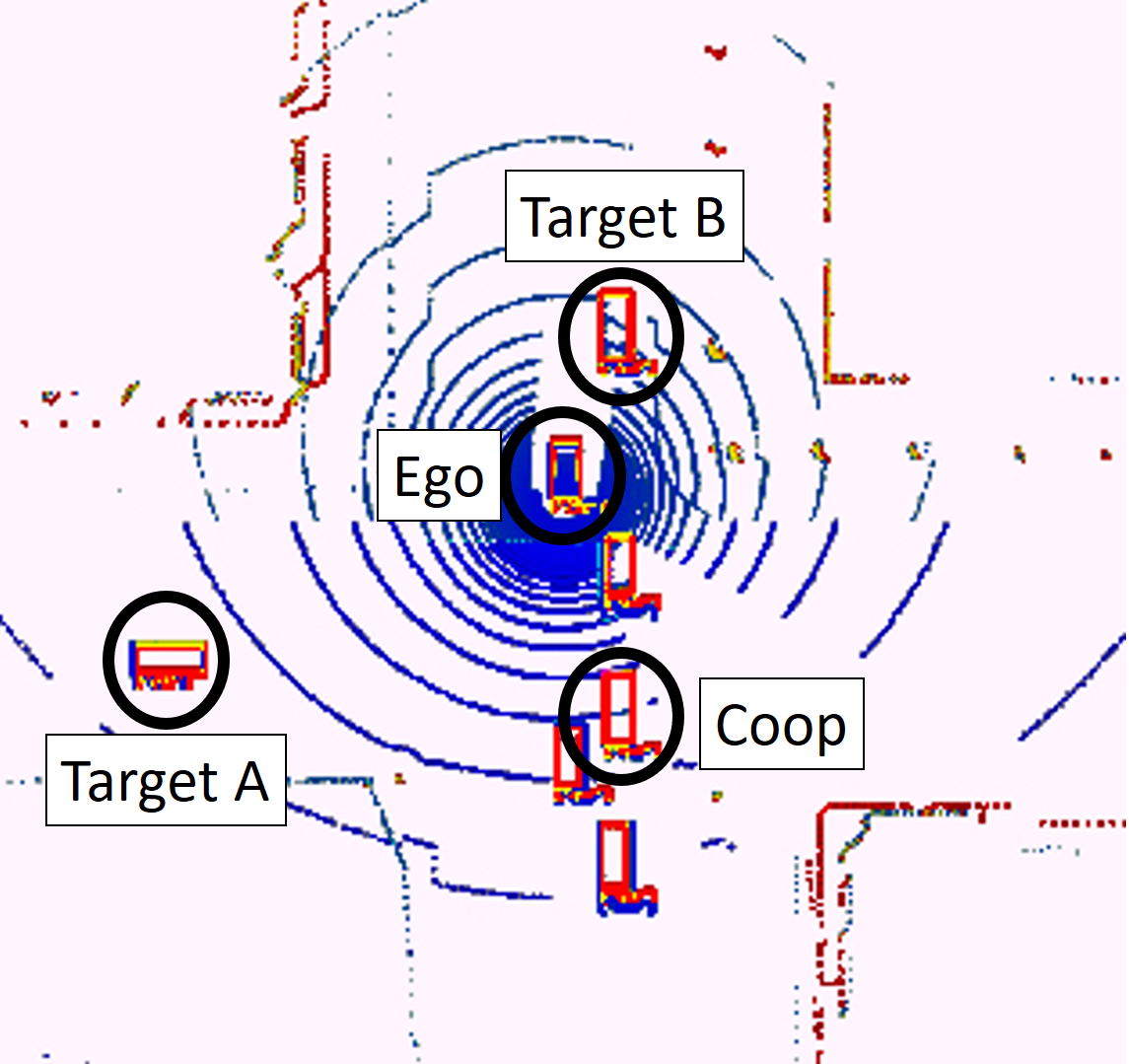}
     }
     \caption{Comparison between performance of single vehicle object detection and feature sharing cooperative object detection in an arbitrary scenario. blue and red bounding boxes represent ground truth and output of the object detection; (a)The single vehicle object detection at ego-vehicle, (b) The single vehicle object detection at coop-vehicle and (c) FS-COD at ego-vehicle}
     \label{fig:FS-COD-BEV-RES}
   \end{figure*}

Fig. \ref{fig:FS-COD-BEV-RES} demonstrates a scenario in which, target $A$ is not detectable by either vehicles and there is a lack of consensus on target $B$ between cooperative vehicles if they rely solely on their own sensory and inference units. However, target A is detectable if FS-COD is applied and the lack of consensus on target $B$ is solved.

   \begin{figure}[!pt]
     \subfloat[]{%
       \includegraphics[width=0.1559\textwidth]{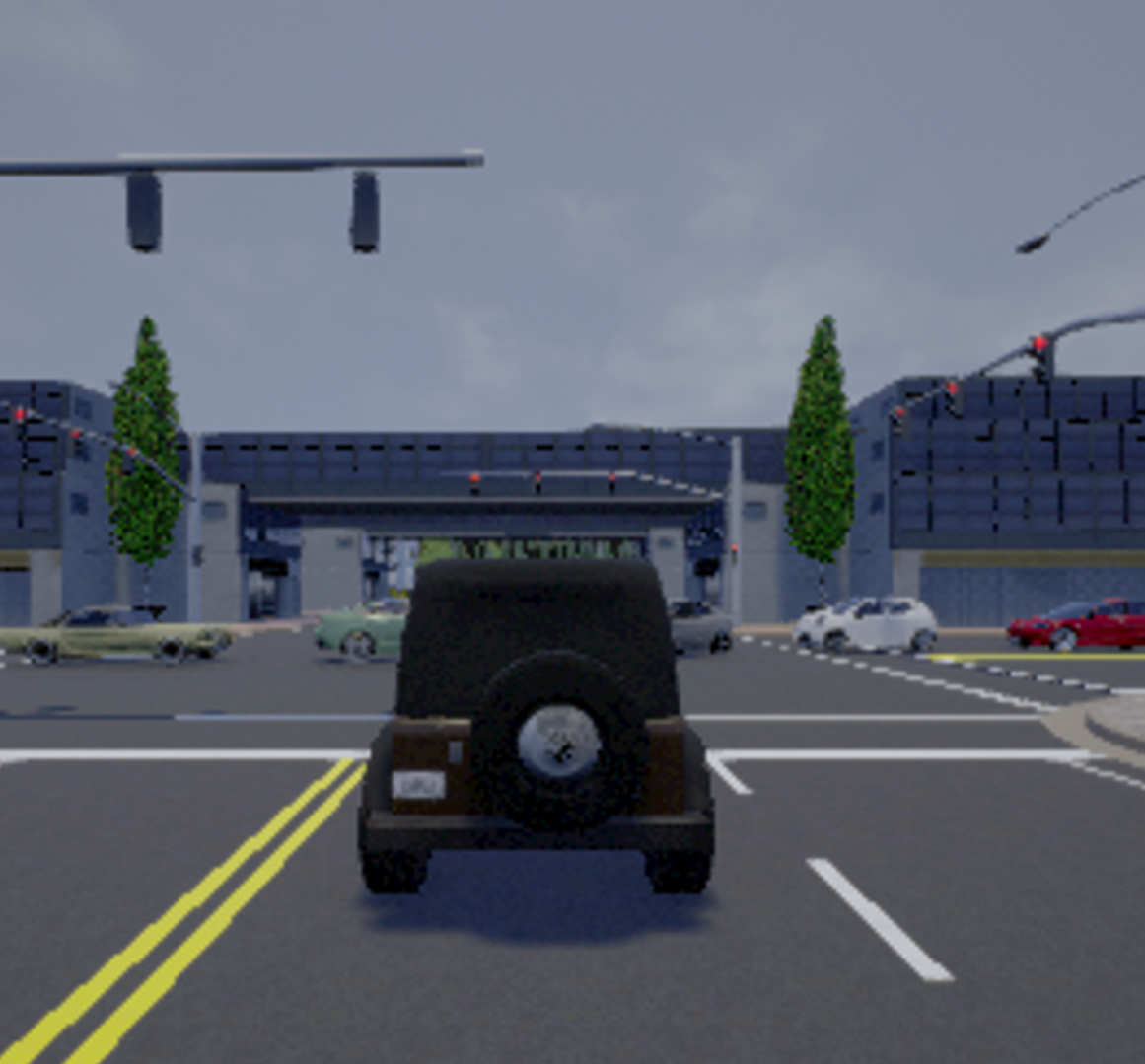}
     }
     \hfill
     \subfloat[]{%
       \includegraphics[width=0.159\textwidth]{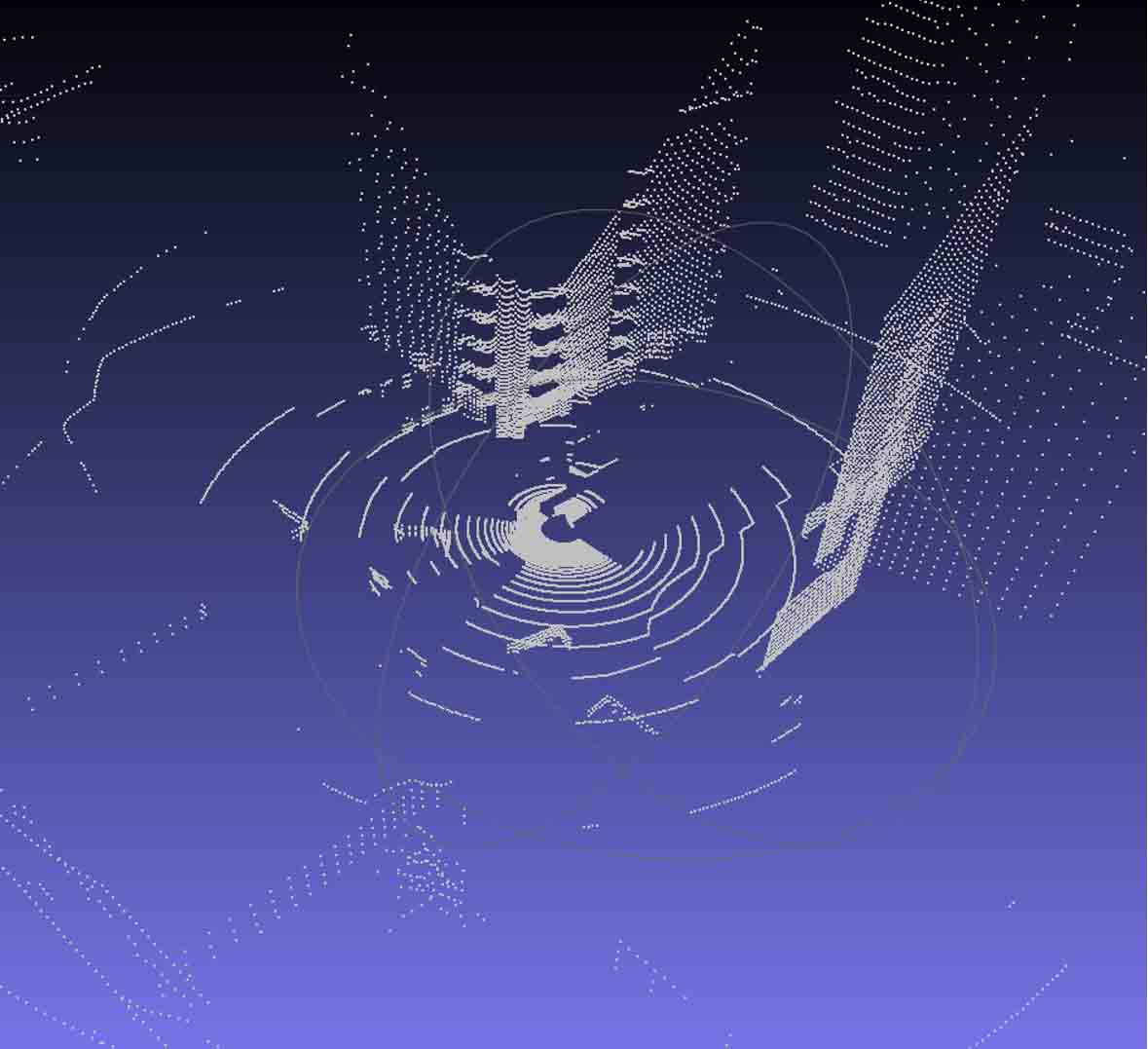}
     }
     \hfill
     \subfloat[]{%
       \includegraphics[width=0.148\textwidth]{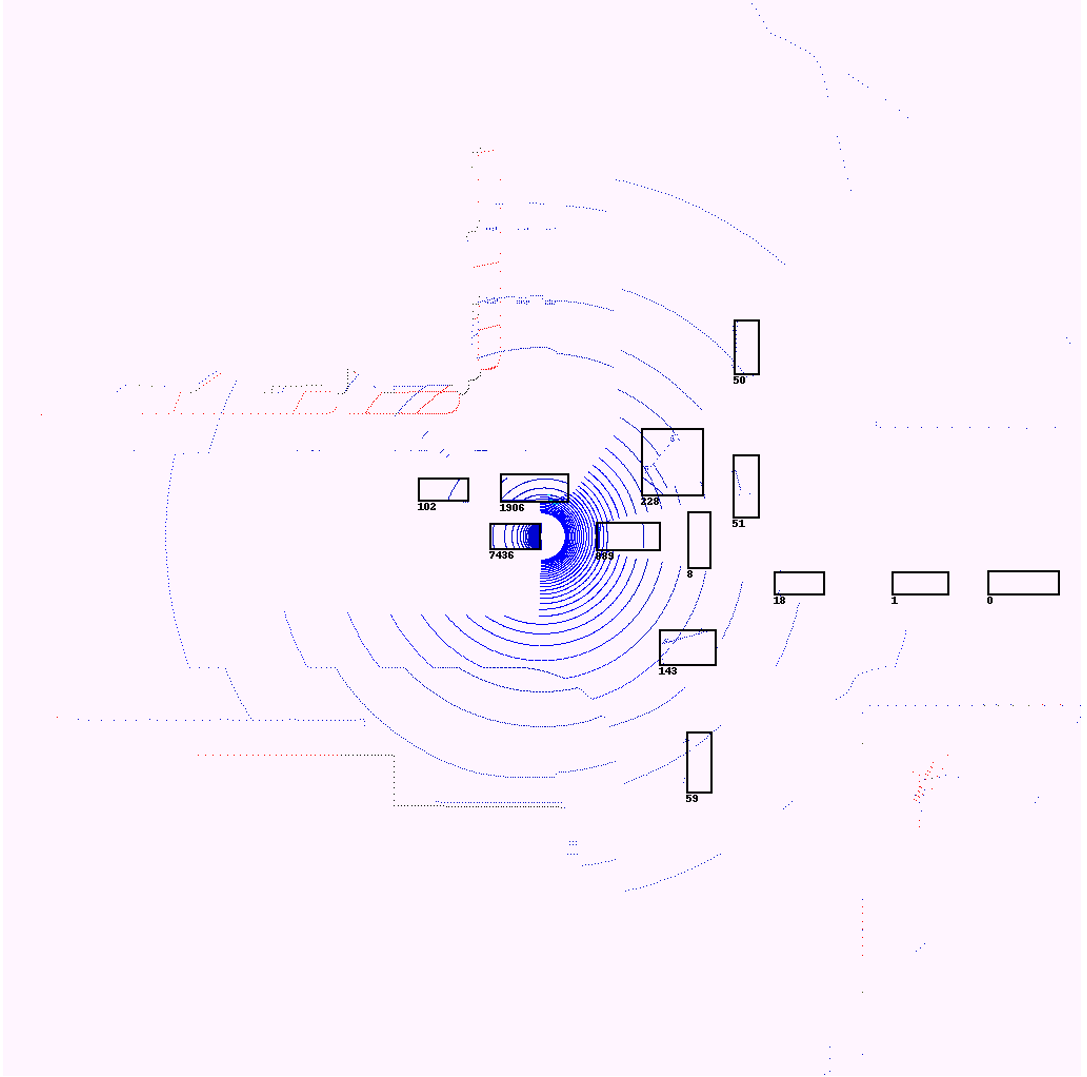}
     }
     \caption{(a) Illustration of CARLA work-space in an arbitrary scenario (b) Point-cloud generated from LIDAR device of a cooperative vehicle (c) BEV representation of the point-cloud}
     \label{fig:foobar}
   \end{figure}



\subsection{FS-COD Architecture}

In this section, we present the detail of FS-COD architecture along with its training procedure and input data representation details. Fig. \ref{fig:overall} illustrates an overview of FS-COD system.

In our proposed setup, each cooperative vehicle is equipped with LIDAR. Point-clouds generated from LIDAR devices are used as observations provided for the detection systems. Additionally, cooperative vehicles share the partially processed information along the metadata containing their positioning information. 
The FS-COD architecture for each participating cooperative vehicle includes BEV projector function, feature extractor CNN structure, feature accumulator and an object detection CNN module along with an alignment procedure. The remainder of this section describes the FS-COD procedure in a sequential manner.

The first step in FS-COD is to align the point-clouds obtained by LIDAR device with respect to vehicles' heading and a predefined global coordinate system.
The alignment is done by considering a global coordinate system for all cooperative entities and rotate vehicles' point clouds represented in local ego coordinate systems with respect to the global coordinate system.
The rotation alignment formulation is as follows.  
\begin{equation}
    X_w = X_e R_x(\alpha)R_y(\beta)R_z(\gamma)
\end{equation}

Where $X_w$ and $X_e$ are representations of a point in the global and local coordinate systems respectively. $R_x$, $R_y$ and $R_z$ are the rotation matrices for $x$, $y$ and $z$ axis.

Subsequently, BEV projector unit projects the aligned LIDAR point-clouds onto 2D image plane. The projection procedure is similar to the projection method proposed in \cite{simony2018complex} with a slight difference. In our projection method, the BEV image has 3 channels and each channel provides the density of reflected points at a specific height bin. The height bins are defined as $[-\infty,2m]$, $[2m,4m]$ and $[4m,\infty]$.

After the projection step, the constructed BEV image is fed into the feature extractor network to acquire the feature-map of the surrounding environment. The feature extractor CNN architecture is shown in Table \ref{archtab}. 

The produced feature-maps are transmitted along with the cooperative vehicle GPS information to other cooperative vehicles. The number of filters at the last layer of the feature extractor ($C_t$ in Table \ref{archtab}) determines the size of data being shared between cooperative vehicles using our proposed approach. Therefore, the bandwidth requirement can be met by customizing the structure of feature extractor CNN and more specifically by tuning the filters at the last convolutional layer. 

All the procedures mentioned above also occurs at the receiver vehicle yielding the feature-map of receiver's projected point-cloud. In the rest of the paper, we refer to the vehicle receiving feature-maps as ego-vehicle and the vehicle transmitting feature-maps as coop-vehicle. 
The received coop-vehicle's feature-map is further aligned with respect to the ego-vehicle local coordinate system and accumulated with ego-vehicle's feature-map.
Since the rotation alignment has already taken place before transmitting, the second phase of alignment is a 2D image translation transformation. The equations for translation alignment are as follows.
\begin{equation}
    \hat{F}_c(x_f,y_f) = F_c(x_f+\Delta x , y_f +\Delta y)
\end{equation}
\begin{equation}
    \Delta x = \floor*{\frac{x_e}{s}}- \floor*{\frac{x_c}{s}}
\end{equation}
\begin{equation}
    \Delta y = \floor*{\frac{y_e}{s}}- \floor*{\frac{y_c}{s}}
\end{equation}

Where $F_f$, $\hat{F}_f$, $(x_c,y_c)$, $(x_e,y_e)$ are the coop-vehicle's feature-map, aligned coop-vehicle's feature-map, coop-vehicle and ego-vehicle pixel-wise locations in global coordinate system respectively. The down-sampling rate from BEV image to feature-map is denoted by $s$. this rate is defined by total number of maxpool layers in the architecture.

Subsequently, the accumulation is done by an element-wise summation of ego-vehicle's and aligned coop-vehicle's feature-maps. We have assumed that the information acquired from ego-vehicle and coop-vehicle has the same level of importance. Therefore, any accumulation function should follow symmetric property with regards to inputs. Based on this assumption, swapping the observation of the transmitter and receiver should not affect the output of object detection system.

Finally, the resulting accumulated feature-map is fed into the object detection CNN module to detect the targets in the environment. The architecture of object detection module is provided in Table \ref{archtab}.

\begin{table}[t]
\caption{The architecture of Proposed networks}
\label{archtab}
\begin{center}
\begin{tabular}{ |c| c|| c| c|}

\hline
\multicolumn{2}{|c||}{10.4 ppm} &\multicolumn{2}{|c|}{4.16 ppm}  \\ \hline
 \textbf{Baseline}&\textbf{FS-COD}&\textbf{Baseline}&\textbf{FS-COD} \\
\hline
\multicolumn{4}{|c|}{Input 832x832x3 } \\ \hline \hline
\multicolumn{4}{|c|}{\textbf{Feature Extraction Component}} \\ \hline
\multicolumn{4}{|c|}{3x3x24 Convolution Batch-Norm Leaky ReLU(0.1)} \\ \hline
\multicolumn{4}{|c|}{Maxpool/2} \\ \hline
\multicolumn{4}{|c|}{3x3x48 Convolution Batch-Norm Leaky ReLU(0.1)}\\ \hline
\multicolumn{4}{|c|}{Maxpool/2} \\ \hline
\multicolumn{4}{|c|}{3x3x64 Convolution Batch-Norm Leaky ReLU(0.1)} \\ \hline
\multicolumn{4}{|c|}{3x3x32 Convolution Batch-Norm Leaky ReLU(0.1)} \\ \hline
\multicolumn{4}{|c|}{3x3x64 Convolution Batch-Norm Leaky ReLU(0.1)} \\ \hline
\multicolumn{4}{|c|}{Maxpool/2} \\ \hline
\multicolumn{4}{|c|}{3x3x128 Convolution Batch-Norm Leaky ReLU(0.1)} \\ \hline
\multicolumn{4}{|c|}{3x3x64 Convolution Batch-Norm Leaky ReLU(0.1)} \\ \hline
\multicolumn{4}{|c|}{3x3x128 Convolution Batch-Norm Leaky ReLU(0.1)} \\ \hline
\multicolumn{2}{|c||}{Maxpool/2}&\multicolumn{2}{|c|}{-} \\ \hline
\multicolumn{4}{|c|}{3x3x128 Convolution Batch-Norm Leaky ReLU(0.1)} \\ \hline
1x1x64&1x1x$C_t$&1x1x64&1x1x$C_t$ \\ \hline \hline 
\multicolumn{4}{|c|}{\textbf{Object Detection Component}} \\ \hline \hline
\multicolumn{4}{|c|}{1x1x128 Convolution Batch-Norm Leaky ReLU(0.1)}\\ \hline
\multicolumn{4}{|c|}{3x3x256 Convolution Batch-Norm Leaky ReLU(0.1)}\\ \hline
\multicolumn{4}{|c|}{1x1x512 Convolution Batch-Norm Leaky ReLU(0.1)}\\ \hline
\multicolumn{4}{|c|}{1x1x1024 Convolution Batch-Norm Leaky ReLU(0.1)}\\ \hline
\multicolumn{4}{|c|}{3x3x2048 Convolution Batch-Norm Leaky ReLU(0.1)}\\ \hline
\multicolumn{4}{|c|}{1x1x1024 Convolution Batch-Norm Leaky ReLU(0.1)} \\ \hline
\multicolumn{4}{|c|}{1x1x2048 Convolution Batch-Norm Leaky ReLU(0.1)}\\ \hline
\multicolumn{4}{|c|}{3x3x1024 Convolution Batch-Norm Leaky ReLU(0.1)} \\ \hline
\multicolumn{4}{|c|}{1x1x20 Convolution}\\ \hline
\multicolumn{2}{|c||}{Output 52x52x20}&\multicolumn{2}{|c|}{Output 104x104x20}\\ \hline
\end{tabular}
\end{center}

\end{table}
\subsection{FS-COD Training Method}
In the previous section, the feed-forward process of FS-COD was discussed. As it was mentioned, the system contains two sets of networks with identical structure, one residing at coop-vehicle and one at ego-vehicle. Here, we briefly explain the technique used for training these networks. The symmetric property for feature accumulation imposes the networks at both vehicles to have identical parameters. 
For training, a single feature extractor network is fed-forward with both vehicles observations. Therefore, the gradients in the back-propagation step are calculated with respect to both observations and the weights of the feature extractor network are updated accordingly. At the next feed-forward step, the same updated network is used for both vehicles. Assuming $g$ to be the feature accumulation function and $f$ to be the feature extractor function, function $g$ is defined as
\begin{equation}
    g(f(Z_1;\theta),f(Z_2;\theta)) = f(Z_1;\theta) + f(Z_2;\theta),
\end{equation}
Where $Z_1$ and $Z_2$ are cooperative vehicles observations and $\theta$ is the feature extractor component parameters. Hence, the partial derivative with respect to shared parameters $\theta$ is calculated by
\begin{equation}
    \frac{\partial g(f(Z_1;\theta),f(Z_2;\theta))}{\partial \theta} = \frac{\partial f(Z_1;\theta)}{\partial \theta} + \frac{\partial f(Z_2;\theta)}{\partial \theta}.
    \label{eq::partial_derivative}
\end{equation}
Equation (\ref{eq::partial_derivative}) can be used in chain rule in order to perform back propagation. The details on loss function can be found in \cite{Redmon_2016_CVPR}.

\section{Experiments and Results}
\label{Section:Experiments and Results}
In this section we provide the details of the dataset on which we have tested FS-COD. Furthermore, the baseline network has been elaborated. The section is concluded with results and evaluation.

\begin{figure*}[t!]
    \centering
    \includegraphics[width=.88\linewidth,trim={30mm 0mm 30mm 0mm},clip]{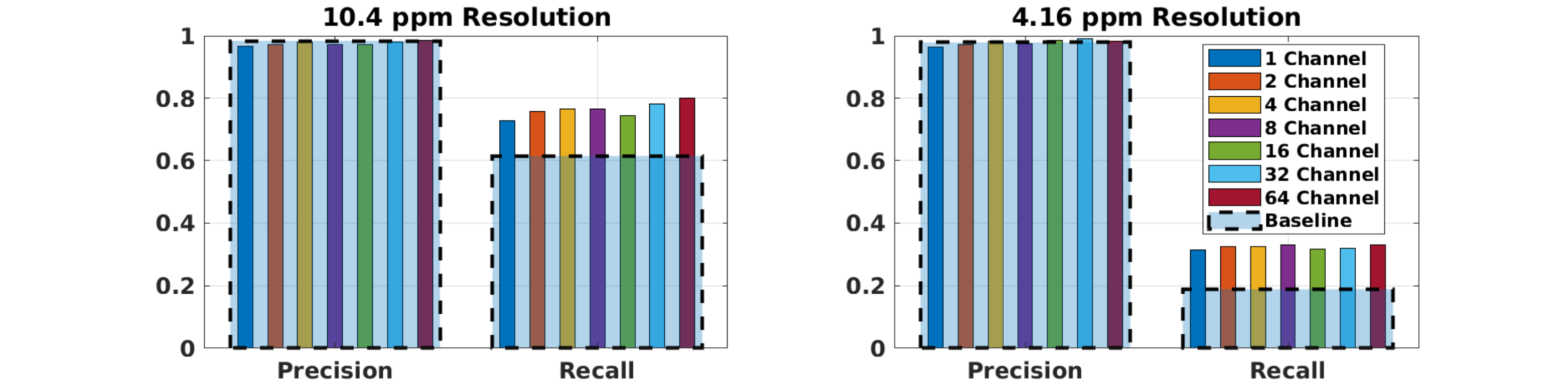}
    \caption{Detection precision and recall of baseline and FS-COD with different feature-map channel sizes. The left figure shows the results of 10.4 ppm resolution (40 meter range LIDAR) experiment while the right figure shows the results of experiment with 4.16 ppm resolution (100 meter range LIDAR}
    \label{fig:precision_recall}
\end{figure*}
\begin{figure*}[t!]
    \centering
    \includegraphics[width=.9\linewidth,trim={20mm 0mm 30mm 0mm},clip]{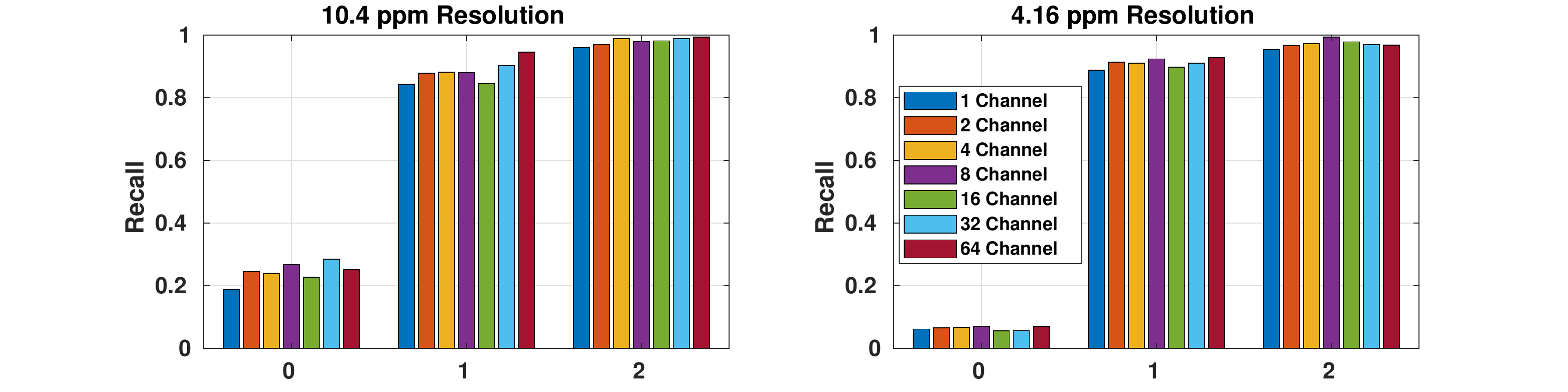}
    \caption{Detection recall of FS-COD with different feature-map channel sizes for different categories. Each category determines how many cooperative vehicle would be able to detect the target if they used the baseline method}
    \label{fig:category_recall}
\end{figure*}
\subsection{Data Gathering and Simulation}
\label{Data Gathering and Simulation}
To the best of our knowledge, no dataset exists to provide simultaneous observations from the same scene for training networks in a cooperative setting.
While this is a significant challenge for research on this subject, we have created the first example of such datasets using simulation tools. We have developed a dataset collection tool called Volony\cite{volony2020} based on CARLA\cite{dosovitskiy2017carla} to test the performance of our proposed method. Using Volony, the user can gather measurements such as RGBD images and LIDAR point-clouds data acquired simultaneously from cooperative vehicles. Additionally, the user has access to objects' bounding box information, labels and metadata such as vehicles' GPS information. Moreover, the characteristics of sensors can be customized and various sensors with different range and precision can be utilized.

To obtain realistic observations, we have deployed vehicles in an urban area containing objects such as buildings, trees, traffic lights.
The urban environment will lead to realistic scenarios in which the observer faces complications like occlusion. 
In our experiments, we deployed 90 vehicles in an urban area in the simulator and equipped 30 vehicles capturing synchronized LIDAR observations from the environment. The measurements are done periodically within 0.1 second intervals; in addition, every 10 seconds the vehicles are redeployed to minimize the correlation between samples. In this paper, to demonstrate the effect of image resolution on the performance of FS-COD, we gathered two different datasets from the simulator. In the first one, the vehicles are equipped with a 40m range LIDAR device and in the second dataset, the range is increased to 100m. In both cases, the size of the 2D image resulting from BEV projection is fixed at 832$\times$832 pixels. Therefore, the image resolution is 10.4 pixels per meter(ppm) and 4.16 ppm respectively. To attain fair comparison, the same set of samples were used for training of both FS-COD and baseline schemes. The selected dataset includes only frames in which at least two cooperative vehicles have measurements from one or more targets in the scene.
For both resolutions, the datasets consist of 5000 and 1000 sample frames for training and validation respectively. The setup of vehicles has been randomly reinitialized for the validation dataset to remove dependency between training and validation sets.

 Fig. \ref{fig:foobar} illustrates the captured observations within CARLA simulator in the urban area setting; captured point-clouds from the LIDAR device are illustrated along with the corresponding BEV image and RGB camera image.

\subsection{Baseline: Single Vehicle Object Detection}
\label{Baseline: Single Vehicle Object Detection}

\begin{figure*}[!pt]
\centering
     \subfloat[Recall]{%
       \includegraphics[width=.87\linewidth,trim={20mm 0mm 30mm 0mm},clip]{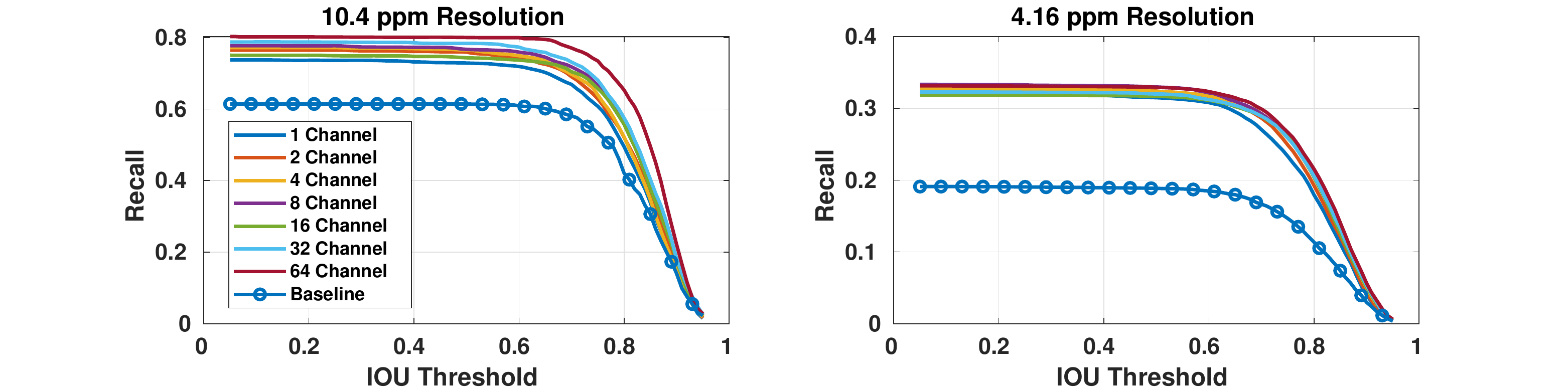}
     }
     \qquad
     \subfloat[Precision]{%
       \includegraphics[width=.87\linewidth,trim={20mm 0mm 30mm 7mm},clip]{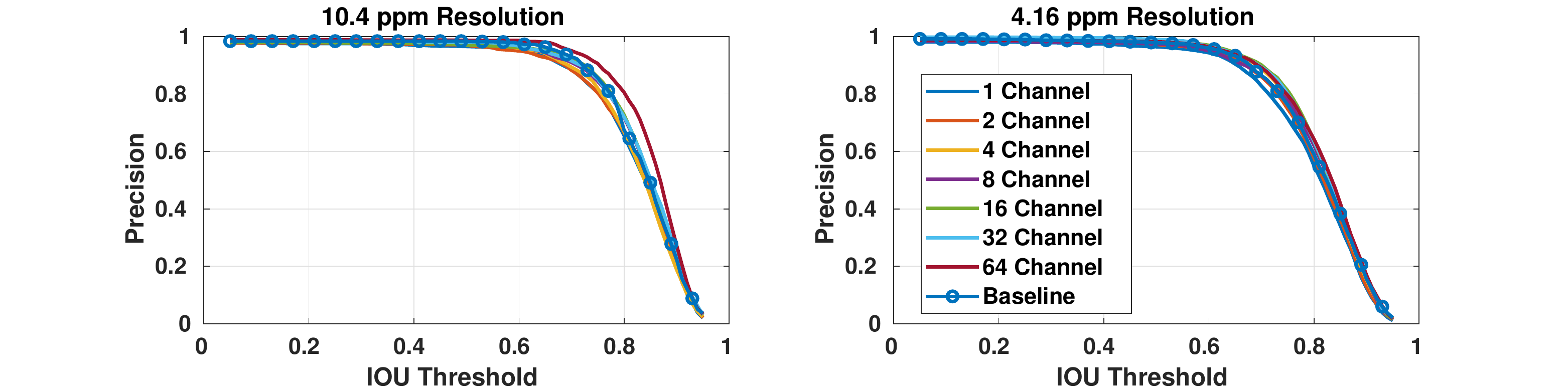}
     }
     \caption{Precision and recall of baseline and FS-COD with different feature-map channel sizes versus different choices of IoU threshold for both experiments with 10.4 ppm and 4.16 ppm resolutions}
     \label{fig::PR-IOU}
   \end{figure*}

 

We designed an individual vehicle's object detection method, based on the data from its on-board sensors to be considered as the baseline object detection method. The feature sharing method is developed out of this baseline architecture. When information shared by other vehicles is not available, the proposed solution is automatically reduced to the baseline method. 

The design of baseline network architecture is adapted and modified from Complex-YOLO architecture. As it was mentioned in previous section, in vehicular applications, some assumptions about the characteristic of the objects can be made, e.g., vehicles are on surfaces and have pre-known sizes with relatively small variations based on the class of vehicle. We exploited these assumptions by modifying Complex-YOLO network architecture to enhance detection performance. This enhancement has been done by carefully choosing the number of maxpool layers, with respect to the size of vehicles (target objects) and input resolution. The total number of maxpool layers is determined such that the output cell's size is approximately equal to or smaller than the width of vehicles in the image. Hence, each cell can only be associated with one object in the training phase. 
Since we have done experiments with two datasets of 4.16 and 10.4 ppm resolutions, two baseline single-class object detector architectures have been designed. 

The details of the baseline network architectures is provided in Table \ref{archtab}. As it is seen, cascading the feature extraction and object detection components of FS-COD design will result in baseline architecture. This property makes both FS-COD and baseline results comparable since both architecture have the same number of parameters.

\subsection{Evaluation and Results}
\label{Evaluation and Results}
In this section, the evaluation of our proposed architecture on the dataset is presented.
Precision and recall, as two commonly used metrics in literature, are opted for assessing object detection performance.

Additionally, we have considered the communication load as a metric to evaluate the effectiveness of FS-COD with different feature extraction networks. 
In reality, the communication channel has limited capacity. Feature sharing concept relies on both distributed processing and observation compression. By transmitting features instead of raw observation, an image with size of $(H,W,C)$ is compressed into a tensor of size $(H_f, W_f,C_t)$. 
Therefore, the communication load is proportionate to the total number of channels.
By enforcing $H_f.W_f.C_t < H.W.C$, we have ensured that the network requires less bandwidth for transmission. 
In our experiments, we evaluate the precision and recall by changing the number of transmitted feature-map channels ($C_t$).

Table \ref{table::bandwidth} shows the shared feature-maps data size with respect to number of channels. We have assumed every element of the map is represented by a 32 bit single-precision floating point. The data size per frame is calculated without the consideration of sparsity in feature-maps. Therefore, the values provided are the upper-bound of data size for a frame.

Fig. \ref{fig:precision_recall} provides the comparison of detection precision and recall for baseline and FS-COD with different feature-map channel sizes for both 10.4 ppm and 4.16 ppm resolutions. In all cases, the detection IOU and object confidence threshold are $50\%$ and $40\%$ respectively. The reported recall is calculated based on all existing targets within the LIDAR range of the ego vehicle, i.e. 40 m or 100 m. Therefore, the targets that are fully occluded are considered in the calculation of recall and precision. The results show FS-COD improves recall significantly while maintaining the precision for both resolution.
As it is seen, FS-COD differentiates itself more from the baseline in 4.16 ppm input resolution experiments. 
The rationale behind such an observation comes from the fact that there are more targets suffering from occlusion in 100m range.

Fig. \ref{fig:category_recall} illustrates a more in-depth approach to observe how object detection benefits from FS-COD. The figure shows the recall of FS-COD in different scenarios based on baseline detection output. The targets in the validation data have been partitioned into three categories based on baseline performance. Each category determines how many cooperative vehicles would be able to detect the target if they used the baseline method. In other words, the members of category $0$, $1$ and $2$ are targets in the area that are detectable by neither cooperative vehicles, exclusively one cooperative vehicle and both cooperative vehicles using baseline respectively. We can observe that on average, FS-COD can detect $30\%$ and $8\%$ of objects where both vehicles are unable to detect in 4.16 ppm and 10.4 ppm resolutions respectively. It should be noted that the majority of targets in category $0$, specifically in experiment with lower resolution, were observable by neither cooperative vehicles, i.e. no reflections from the target were received by vehicles. Additionally, in category 1, FS-COD has remedied more than $80\%$ of situations where cooperative vehicles had no consensus on their inferences. Finally, category 2 results guarantee that FS-COD does not have detrimental effect on object detection in more than $97\%$ of the situations.

Fig. \ref{fig::PR-IOU} illustrates detection precision and recall of baseline and FS-COD with different feature-map channel sizes versus Intersection-over-Union (IoU) threshold. We have incrementally increased the IoU threshold to compare the localization performance of FS-COD with baseline. As it is clearly demonstrated, FS-COD maintains its superior performance even at very large IoU thresholds compared to baseline.

\begin{table}[t]
\caption{Shared Features Data Size}
\label{table::bandwidth}
\begin{center}
\begin{tabular}{ |c|c|c|}

\hline
 \textbf{TX-Channel}&\textbf{10.4 ppm resolution}&\textbf{4.16 ppm resolution} \\
\hline
1&10KB&40KB \\ \hline 
2&20KB&80KB \\ \hline 
4&40KB&160KB \\ \hline
8&80KB&320KB \\ \hline 
16&160KB&640KB \\ \hline
32&320KB&~1MB \\ \hline
64&640KB&~2MB \\ \hline
\end{tabular}
\end{center}
\end{table}

\section{Concluding Remarks and Future work}
\label{Section:Concluding Remarks and Future work}

We have proposed a novel framework to improve object detection by integrating communication and CNN based inference systems. The proposed cooperative method, FS-COD, relies on sharing of extracted features from sensory data (e.g., LIDAR data). By enabling the cooperative participants to effectively compress and encode relevant information, FS-COD remedies the inherent limitations of sensory and communication systems while maintaining the object detection performance at a desirable level.
We have shown that by transmitting 10KB for a point cloud frame, we can improve recall significantly for single class object detection while maintaining the precision.

The promising improvement resulting from utilizing the novel concept of feature sharing motivates us to further investigate its application in other object detection approaches such as point-cloud volumetric and RGB monocular methods. Utilizing such inference methods in cooperative context using feature sharing requires developing new alignment and feature extraction procedures.

\bibliographystyle{IEEEtran}\small
\bibliography{references}

\end{document}